\documentclass[10pt,twocolumn,letterpaper]{article}

\usepackage{cvpr}              

\usepackage{graphicx}
\usepackage{amsmath}
\usepackage{amssymb}
\usepackage{xfrac}
\usepackage{color}
\usepackage{colortbl}
\usepackage{enumitem}
\usepackage[dvipsnames]{xcolor}
\newcommand{\Lagr}{\mathcal{L}}
\usepackage{sidecap}
\usepackage{wrapfig}
\usepackage{lipsum} 
\usepackage{textcmds}
\usepackage{graphicx}
\usepackage{multirow}
\usepackage{booktabs}
\usepackage{colortbl}
\usepackage{xcolor}
\usepackage{hhline}
\usepackage{lipsum}
\usepackage{etoolbox}
\usepackage{capt-of,etoolbox}
\definecolor{Gray}{gray}{0.90}
\definecolor{LightCyan}{rgb}{0.82,0.82,1}
\definecolor{tabhighlight}{HTML}{e5e5e5}

\newcommand{\txt}[1]{{\texttt{#1}}}
\def\etal{\emph{et al.}\xspace}

\newcommand\extrafootertext[1]{%
    \bgroup
    \renewcommand\thefootnote{\fnsymbol{footnote}}%
    \renewcommand\thempfootnote{\fnsymbol{mpfootnote}}%
    \footnotetext[0]{#1}%
    \egroup
}

\usepackage[pagebackref,breaklinks,colorlinks]{hyperref}

\usepackage[capitalize]{cleveref}
\crefname{section}{Sec.}{Secs.}
\Crefname{section}{Section}{Sections}
\Crefname{table}{Table}{Tables}
\crefname{table}{Tab.}{Tabs.}

\def\cvprPaperID{1954} 
\def\confName{CVPR}
\def\confYear{2023}

\begin{document}

\title{Fine-tuned CLIP Models are Efficient Video Learners}


\author{%
  Hanoona Rasheed$^{1*}$ \quad 
  Muhammad Uzair Khattak$^{1*}$ \quad 
  Muhammad Maaz$^{1}$ \\
  Salman Khan$^{1,2}$ \quad
  Fahad Shahbaz Khan$^{1,3}$
  \vspace{0.2em} \\
  $^{1}$Mohamed bin Zayed University of AI \quad 
  $^{2}$Australian National University \quad 
  $^{3}$Link\"{o}ping University
}

\thispagestyle{empty}
\twocolumn[{
    \renewcommand\twocolumn[1][]{#1}
    \vspace{-1em}
    \maketitle
    \vspace{-0.9cm}
    \begin{center}
        \centering
        \captionsetup{type=figure}
        {\includegraphics[width=0.98\textwidth]{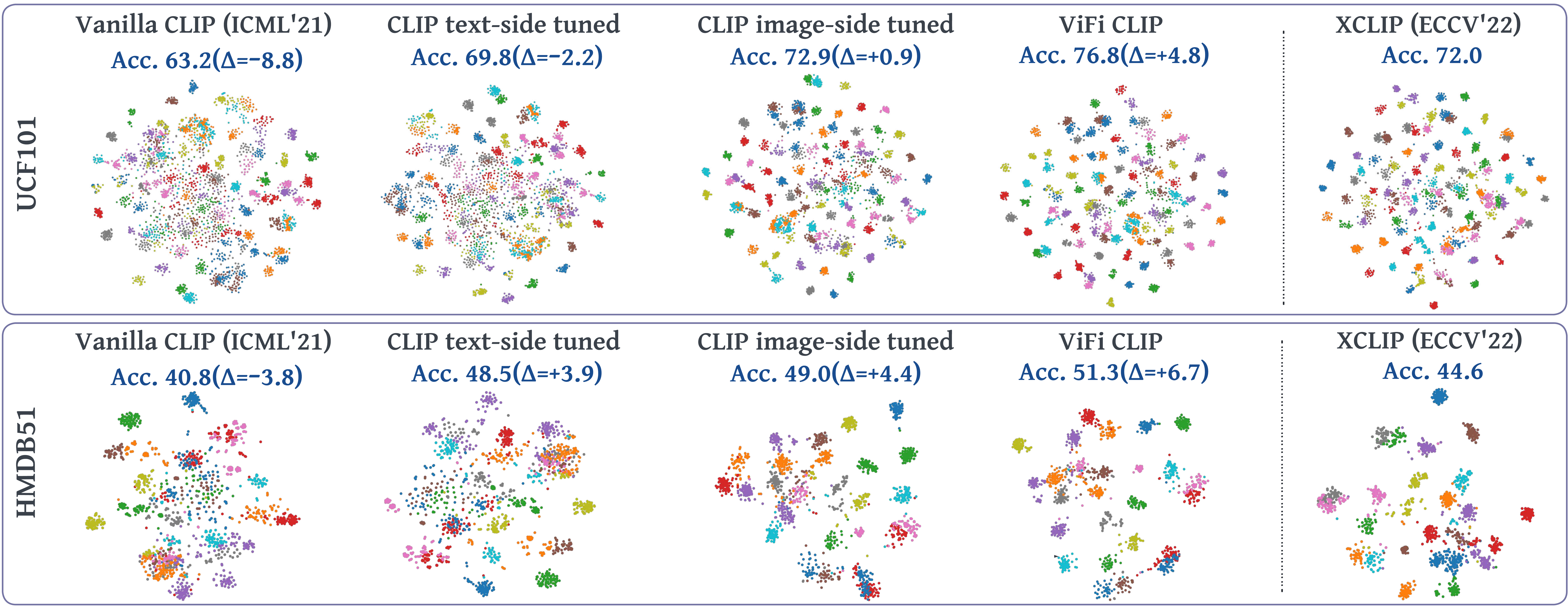}}
        \captionof{figure}{This work explores the capability of a simple baseline called {ViFi-CLIP} ({Video Fintuned CLIP}) for adapting image pretrained CLIP~\cite{clip} to video domain. The figure compares the zero-shot performance of vanilla CLIP and several of its variants adapted for videos (trained on Kinetics-400, evaluated on UCF-101 and HMDB-51). The t-SNE visualizations of video-embeddings obtained from ViFi-CLIP ($4^{th}$ col.) are compared with embeddings from vanilla CLIP~\cite{clip} ($1^{st}$ col.), individually tuned CLIP text ($2^{nd}$ col.) and image encoder ($3^{rd}$ col.) on videos, and recent state-of-the-art work, XCLIP~\cite{ni2022expanding} (last col.) ($\Delta$ represents difference over XCLIP). The embeddings of ViFi-CLIP are better separable, indicating that a simple fine-tuning of CLIP is sufficient to learn suitable video-specific inductive biases, and can perform competitive to more complex approaches having dedicated components designed to model temporal information in videos.}
        \label{fig_1:tsne_plot_into}
    \end{center}
}]

\extrafootertext{\textsuperscript{*}Equally contributing authors.}
\begin{abstract}\vspace{-1em}
Large-scale multi-modal training with image-text pairs imparts strong generalization to CLIP model. Since training on a similar scale for videos is infeasible, recent approaches focus on the effective transfer of image-based CLIP to the video domain. In this pursuit, new parametric modules are added to learn temporal information and inter-frame relationships which require meticulous design efforts. Furthermore, when the resulting models are learned on videos, they tend to overfit on the given task distribution and lack in generalization aspect. This begs the following question: How to effectively transfer image-level CLIP representations to videos? In this work, we show that a simple Video Fine-tuned CLIP (ViFi-CLIP) baseline is generally sufficient to bridge the domain gap from images to videos. Our qualitative analysis illustrates that the frame-level processing from CLIP image-encoder followed by feature pooling and similarity matching with corresponding text embeddings helps in implicitly modeling the temporal cues within ViFi-CLIP. Such fine-tuning helps the model to focus on scene dynamics, moving objects and inter-object relationships. For low-data regimes where full fine-tuning is not viable, we propose a `bridge and prompt' approach that first uses fine-tuning to bridge the domain gap and then learns prompts on language and vision side to adapt CLIP representations. We extensively evaluate this simple yet strong baseline on zero-shot, base-to-novel generalization, few-shot and fully supervised settings across five video benchmarks. Our code and pre-trained models are available at 
\href{https://github.com/muzairkhattak/ViFi-CLIP}{https://github.com/muzairkhattak/ViFi-CLIP}.
\end{abstract}

\section{Introduction}
\label{sec:intro}
\noindent Pretrained vision-language (VL) models like CLIP~\cite{clip} and ALIGN~\cite{jia2021scaling} have shown impressive zero-shot performance for many downstream vision applications including classification~\cite{zhang2021tip, zhou2022conditional}, detection~\cite{rasheed2022bridging, gu2021open, zhou2022detecting} and segmentation~\cite{rao2021denseclip, li2022language}. These models are trained using millions of image-text pairs sourced from the internet and offer unique representations with strong generalization and transfer capabilities. However, such massive pretraining is laborious for videos due to the following reasons: \textbf{1)} Aligned video-text data has a limited availability and the cost of preparing such data is monumental in contrast to image-text pairs that are readily available via internet sources~\cite{jia2021scaling}. \textbf{2)} Videos are inherently complex and have large compute cost while the diverse appearance cues could be learned through image-text pairs with a much lower compute budget. Therefore, it is critical to devise methods effectively adapting pretrained image-language models for video-based tasks without forgetting the generic multi-modal learned representations.

Recent video-based approaches adopt CLIP representations using additional learnable components for spatio-temporal modeling. These components include self-attention layers for cross-frame communication~\cite{ju2022prompting}, textual or visual prompts~\cite{wang2021actionclip} or dedicated video decoder modules~\cite{ni2022expanding} that are learned while keeping the CLIP backbone frozen or adapting the CLIP encoders as well. However, these designs require modality-specific inductive biases to be modeled in the developed architectural modules and need careful design efforts to adapt CLIP suitably for videos. Additionally, while adapting CLIP for  downstream video tasks, such approaches generally do not remain a winner across all settings. For example, zero-shot adapted approaches perform lower in supervised settings, and supervised models score lower on zero-shot generalization tasks.

To address the above challenges, we frame the following two questions: \textbf{1)} Does the adaptation of CLIP for videos using additional tunable parameters tamper its generalization capacity? \textbf{2)} Is a simple video-specific fine-tuning sufficient to bridge the modality gap between images and videos? In our empirical analysis, we observe that fine-tuning pretrained CLIP encoders along with the newly introduced temporal modeling components can hinder the generalization capability of CLIP. Interestingly, a simple CLIP model when fine-tuned on a video dataset can instill suitable video-specific adaptations within the regular CLIP model and perform competitively to more complex approaches having video-specific components inbuilt.

Although existing works explore fine-tuning of CLIP encoders as a baseline, they undermine the potential of full fine-tuning of CLIP. However we note that, full fine-tuning to achieve better visual-language alignment on videos improves synergy between temporal and language cues,
and perform competitive to much sophisticated approaches developed for videos~(see Fig.~\ref{fig_1:tsne_plot_into}).
Towards understanding how this capacity is achieved by the regular CLIP model, we show that a simple frame-level late representation aggregation before loss calculation allows the exchange of temporal cues within the video fine-tuned CLIP.

While simple CLIP fine-tuning performs competitively to more sophisticated approaches, it is not always feasible, especially on low-data regimes. Based on the finding that simple fine-tuning can efficiently adapt CLIP for videos, we propose a two-stage `\emph{bridge and prompt}' approach for adapting CLIP for low-data regimes that first fine-tunes vanilla CLIP on videos to bridge the modality gap, followed by a vision-language prompt learning approach keeping the tuned CLIP frozen. 
The contributions of this work are,
\vspace{-0.8em}
\begin{itemize}\setlength{\itemsep}{-0.5em}
    \item We 
    formulate a simple but strong baseline, ViFi-CLIP (Video Finetuned CLIP),  for adapting  image-based CLIP to video-specific tasks. We show that simple fine-tuning of CLIP is sufficient to learn video-specific inductive biases, resulting in impressive performance on downstream tasks~(Sec.~\ref{baseline}).
    \item We conduct experiments on four different experimental settings including zero-shot, base-to-novel generalization, few-shot and fully-supervised tasks. We show better or competitive performance as compared to the state-of-the-art approaches~(Secs.~\ref{sec:probel_setting}, \ref{baseline}).
    \item We show the effectiveness of our proposed `bridge and prompt' approach to first bridge the modality gap through fine-tuning followed by prompt learning in both visual and language branches of the CLIP model for low-data regimes~(Sec.~\ref{prompting}).
\end{itemize}

\section{Related Work}
\label{sec:related_work}
\noindent
\textbf{Vision Language models:} 
Learning multi-modal representations using large-scale image-text pretraining has proved to be effective for a wide range of uni-modal and multi-modal applications~\cite{OSCAR, UNITER, VisualBERT, ViLBERT, mdetr, mavl}. Foundational VL models like CLIP~\cite{clip} and ALIGN~\cite{jia2021scaling} follow such a pretraining paradigm and are trained on large-scale image-caption pairs with contrastive self-supervised objectives. These models are open vocabulary and effectively transfer on downstream vision applications including few-shot and zero-shot recognition~\cite{zhou2021coop, zhang2021tip, zhou2022conditional}, object detection~\cite{gu2021open, zhou2022detecting, rasheed2022bridging}, and image segmentation~\cite{zhou2021denseclip, li2022language, Ding_2022_CVPR}. However, adapting pretrained VL models to videos is a challenging task due to the lack of video-specific temporal cues in the image-level pretraining. Therefore, recent works~\cite{ni2022expanding, wang2021actionclip, ju2022prompting} adapt CLIP for videos by incorporating additional learnable components such as self-attention layers, textual or vision prompts, or dedicated visual decoder, and have demonstrated improvements on video applications. However, it is still unclear how much benefit these relatively complex domain-specific components provide compared to simple alternatives such as fine-tuning on videos. 

\noindent
\textbf{Video Action Recognition:}
Designing accurate video understanding models inherently requires encoding both spatial and motion cues. Recently, vision transformers based networks~\cite{arnab2021vivit, liu2022video, yan2022multiview} proposes to effectively model the long range spatio-temporal relationships and have shown consistent improvements over the 3D CNNs~\cite{carreira2017quo,christoph2016spatiotemporal, feichtenhofer2019slowfast, wang2017spatiotemporal}. 

While these approaches follow independent uni-modal solutions, works like ActionCLIP~\cite{wang2021actionclip}, XCLIP~\cite{ni2022expanding} and Ju \etal~\cite{ju2022prompting} adopt a multi-modal approach by utilizing CLIP and steer it for video-understanding tasks. These methods utilize the rich generalized VL representations of CLIP and fuse them with additional components for temporal modeling. However, we note that such design choices can affect the generalization ability of CLIP and a simple fine-tuning approach performs competitively.

\noindent
\textbf{Prompt Learning:}
Prompting is a recently adapted paradigm to efficiently transfer a model to downstream tasks without re-learning the trained model parameters. By utilizing a few additional learnable tokens at the inputs, this approach typically aims to retain the model generalization ability in addition to better transferring the model to downstream applications. Inherited from the NLP domain, prompting has been widely used for many vision and V-L models. CoOp~\cite{zhou2021coop} and CoCoOp~\cite{zhou2022conditional} propose to use continuous learnable text prompts to transfer CLIP for image recognition tasks. Bahng~\etal~\cite{bahng2022visual} introduce visual prompts to probe CLIP at its vision branch. MaPLe~\cite{khattak2022MaPLe} propose multi-modal prompting to effectively adapt CLIP. By keeping the original model parameters frozen, prompting is considered efficient and requires less computing and training time as compared to conventional full fine-tuning.

In video tasks, Ju~\etal~\cite{ju2022prompting} adapt CLIP via text prompts and transformer layers for temporal modeling. However, this temporal modeling hinders the CLIP generalization, and struggles to perform well in zero-shot setting.

\begin{SCtable*}[\sidecaptionrelwidth][h!]
\centering
\scalebox{0.8}{
\begin{minipage}{\columnwidth}
\centering
\setlength{\tabcolsep}{4mm}{
\resizebox{0.98\columnwidth}{!}{
\begin{tabular}{lcc} \toprule
    Method  & HMDB-51 & UCF-101 \\ \midrule
    \midrule
    \rowcolor{tabhighlight}\multicolumn{3}{c}{Uni-modal zero-shot action recognition models} \\
    \midrule
    ASR ~\cite{wang2017alternative} & 21.8 $\pm$ 0.9 & 24.4 $\pm$ 1.0 \\
    ZSECOC ~\cite{qin2017zero}  &  22.6 $\pm$ 1.2 & 15.1 $\pm$ 1.7 \\
    UR ~\cite{zhu2018towards} & 24.4 $\pm$ 1.6 & 17.5 $\pm$ 1.6 \\
    E2E ~\cite{brattoli2020rethinking}  & 32.7 & 48 \\
    ER-ZSAR ~\cite{chen2021elaborative}  & 35.3 $\pm$ 4.6 & 51.8 $\pm$ 2.9 \\
    \midrule
    \rowcolor{tabhighlight}\multicolumn{3}{c}{Adapting pre-trained image VL models} \\
    \midrule
    Vanilla CLIP~\cite{clip} & 40.8 $\pm$ 0.3 & 63.2 $\pm$ 0.2 \\
    ActionCLIP~\cite{wang2021actionclip} & 40.8 $\pm$ 5.4 & 58.3 $\pm$ 3.4 \\
    XCLIP~\cite{ni2022expanding} &  \underline{44.6} $\pm$ 5.2 & \underline{72.0} $\pm$ 2.3 \\
    A5~\cite{ju2022prompting} & 44.3 $\pm$ 2.2 & 69.3 $\pm$ 4.2 \\
    \midrule
    \rowcolor{tabhighlight}\multicolumn{3}{c}{Tuning pre-trained image VL models} \\
    \midrule
    CLIP image-FT & 49.0 $\pm$ 0.3 & 72.9 $\pm$ 0.8 \\
    CLIP text-FT  & 48.5 $\pm$ 0.1 & 69.8 $\pm$ 1.1 \\
    ViFi-CLIP       & \textbf{51.3} $\pm$ 0.6 & \textbf{76.8} $\pm$ 0.7 \\
                         & {\textcolor{MidnightBlue}{{+6.7}}} &  {\textcolor{MidnightBlue}{{+4.8}}}\\
\bottomrule
\end{tabular}}
}
\end{minipage}
\noindent\hfill
\hspace{-0.1in}
\begin{minipage}{\columnwidth}
\centering
\setlength{\tabcolsep}{2.5mm}{
\resizebox{0.98\columnwidth}{!}{
        \begin{tabular}{lcc} \toprule
    Method & K600 (Top-1) & K600 (Top-5) \\ \midrule
    \midrule
    \rowcolor{tabhighlight}\multicolumn{3}{c}{Uni-modal zero-shot action recognition models} \\
    \midrule
    SJE~\cite{akata2015evaluation}  & 22.3 $\pm$ 0.6 & 48.2 $\pm$ 0.4 \\
    ESZSL~\cite{romera2015embarrassingly} & 22.9 $\pm$ 1.2 & 48.3 $\pm$ 0.8 \\
    DEM~\cite{zhang2017learning}  & 23.6 $\pm$ 0.7 & 49.5 $\pm$ 0.4 \\
    GCN~\cite{ghosh2020all}  & 22.3 $\pm$ 0.6 & 49.7 $\pm$ 0.6 \\
    ERZSAR~\cite{chen2021elaborative}  & 42.1 $\pm$ 1.4 & 73.1 $\pm$ 0.3 \\
    \midrule
    \rowcolor{tabhighlight}\multicolumn{3}{c}{Adapting pre-trained image VL models} \\
    \midrule
    Vanilla CLIP~\cite{clip} & 59.8 $\pm$ 0.3 & 83.5 $\pm$ 0.2 \\
    ActionCLIP~\cite{wang2021actionclip} &  \underline{66.7} $\pm$ 1.1 & \underline{91.6} $\pm$ 0.3 \\
    XCLIP~\cite{ni2022expanding} & 65.2 $\pm$ 0.4 & 86.1 $\pm$ 0.8 \\
    A5~\cite{ju2022prompting} & 55.8 $\pm$0.7  & 81.4 $\pm$ 0.3 \\
    \midrule
    \rowcolor{tabhighlight}\multicolumn{3}{c}{Tuning pre-trained image VL models} \\
    \midrule
    CLIP image-FT & 62.4 $\pm$ 1.0 & 85.8 $\pm$0.5  \\
    CLIP text-FT  & 68.5 $\pm$ 1.2 & 89.6 $\pm$0.3 \\
    ViFi-CLIP       & \textbf{71.2} $\pm$ 1.0 & \textbf{92.2} $\pm$0.3  \\
    & {\textcolor{MidnightBlue}{{+4.5}}} &  {\textcolor{MidnightBlue}{{+0.6}}}\\
    \bottomrule
    \end{tabular}
    }
} 
\end{minipage}}
\hspace{-0.1in}
\caption{
\textbf{Zero-shot setting:}
We compare ViFi-CLIP with uni-modal methods specifically designed for zero-shot action recognition and methods that explicitly adapt CLIP for videos. Models are trained on Kinetics-400 and evaluated directly on HMDB-51, UCF-101 (\textcolor{blue}{left}) and Kinetics-600 (\textcolor{blue}{right}). ViFi-CLIP acheives strong generalization. Accuracy gains over prior best are indicated in \textcolor{MidnightBlue}{blue}. We \underline{underline} the second best numbers.
}
\vspace{-0.1in}
\label{p1_zeroshot}
\end{SCtable*}
\begin{SCtable*}[][!ht]
\setlength{\tabcolsep}{5.25pt}
\scalebox{0.81}{
\begin{tabular}[t]{l ccc|ccc|ccc|ccc}
\toprule
&      \multicolumn{3}{c}{K-400}    &\multicolumn{3}{c}{HMDB-51}  &\multicolumn{3}{c}{UCF-101}    & \multicolumn{3}{c}{SSv2} \\
\cmidrule(lr){2-13}
Method  & Base & Novel & HM          & Base & Novel & HM         & Base & Novel & HM         & Base & Novel & HM   \\
\midrule
\rowcolor{tabhighlight}\multicolumn{13}{c}{Adapting pre-trained image VL models} \\
\midrule
Vanilla CLIP~\cite{clip}      & 62.3 & 53.4 & 57.5        & 53.3 &  \underline{46.8} & 49.8       & 78.5 & \underline{63.6} & 70.3       & 4.9 & 5.3 & 5.1 \\ 
ActionCLIP~\cite{wang2021actionclip}        & 61.0 & 46.2 & 52.6         & 69.1 & 37.3 & 48.5        & 90.1 & 58.1 & 70.7       &  \underline{13.3} &  \underline{10.1} &  \underline{11.5} \\
XCLIP~\cite{ni2022expanding}              &  \underline{74.1} &  \underline{56.4} &  \underline{64.0}        &  \underline{69.4} & 45.5 &  \underline{55.0}     & 89.9 &  58.9 &  \underline{71.2}       & 8.5 & 6.6 & 7.4 \\
A5~\cite{ju2022prompting}     & 69.7 & 37.6 & 48.8        & 46.2 & 16.0 & 23.8        &  \underline{90.5} & 40.4 & 55.8       & 8.3 & 5.3 & 6.4\\
\midrule
\rowcolor{tabhighlight}\multicolumn{13}{c}{Tuning pre-trained image VL models} \\
\midrule
CLIP image-FT     & 72.9 & 58.0 & 64.6         & 62.6 & 47.5 & 54.0        & 86.4 & 65.3 & 74.4      & 9.2 & 8.5 & 8.8 \\ 
CLIP text-FT     & 73.4 & 59.7 & 65.8         & 70.0 & 51.2 & 59.1        & 90.9 & 67.4 & 77.4       & 12.4 & 9.5 & 10.8 \\ 
ViFi-CLIP     & \textbf{76.4} & \textbf{61.1} & \textbf{67.9}         & \textbf{73.8} & \textbf{53.3} & \textbf{61.9}        & \textbf{92.9} &\textbf{67.7} & \textbf{78.3}       & \textbf{16.2} & \textbf{12.1} & \textbf{13.9}\\
                  &  \textcolor{MidnightBlue}{{+2.3}} &  \textcolor{MidnightBlue}{{+4.7}}  &  \textcolor{MidnightBlue}{{+3.9}}  &  \textcolor{MidnightBlue}{{+4.4}}  &  \textcolor{MidnightBlue}{{+6.5}}  &  \textcolor{MidnightBlue}{{+6.9}}  &  \textcolor{MidnightBlue}{{+2.4}}  &  \textcolor{MidnightBlue}{{+4.1}}  &  \textcolor{MidnightBlue}{{+7.1}}  &  \textcolor{MidnightBlue}{{+2.9}}  &  \textcolor{MidnightBlue}{{+2.0}}  &  \textcolor{MidnightBlue}{{+2.4}} \\ \bottomrule
\end{tabular}
}
\caption{\textbf{Base-to-novel generalization:} We compare the generalization ability of ViFi-CLIP with models that adapt CLIP~\cite{clip} for video tasks on Kinetics-400, HMDB-51, UCF-101 and SSv2. Here, HM refers to harmonic mean which measures the trade-off between base and novel accuracy. Gains over prior best are shown in \textcolor{MidnightBlue}{blue}.}
\label{tab2}
\end{SCtable*}

\section{Problem Settings}
\label{sec:probel_setting}
\noindent In this section, we introduce four problem settings for video recognition by varying the level of supervision available. This allows us to analyse the performance of our baseline and its comparison with state-of-the-art approaches across a spectrum of tasks with different degrees of generalization required. Below, we discuss the studied settings, including a newly proposed base-to-novel generalization setting for videos, in the increasing order of supervision available.

\noindent
\textbf{Zero-shot setting}: The model is trained on a source dataset and transferred directly on downstream cross-datasets. The source dataset $D_S$ contains samples belonging to source classes, $Y_S=\{y_i\}_{i=0}^{k}$. The model is evaluated on the target dataset $D_T$ with classes $Y_T$ such that $Y_S\cap Y_T=\phi$.

\begin{figure}[!t]
\centering
{\includegraphics[width=0.8\columnwidth]{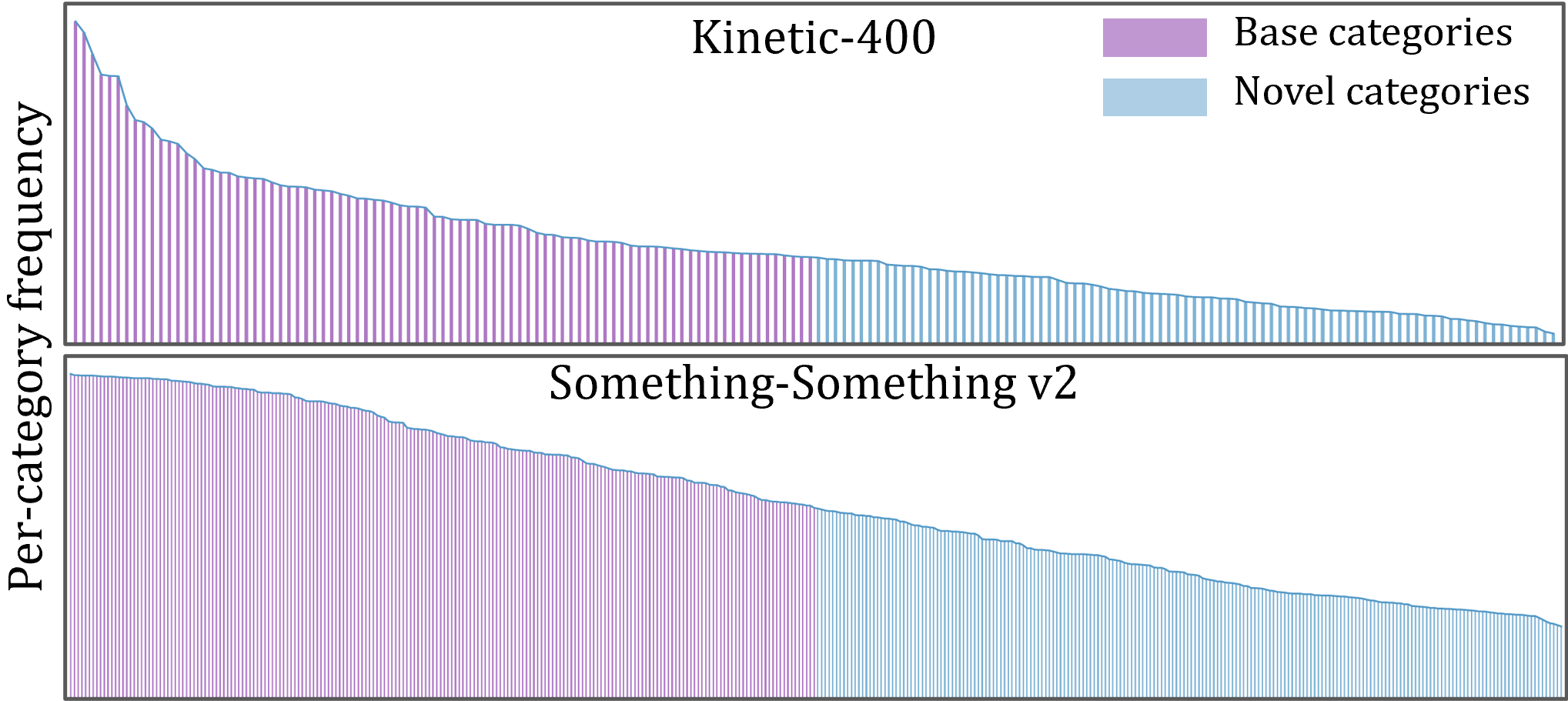}}
\caption{Frequency plot of K400~\cite{k400} and SSv2~\cite{goyal2017ssv2}.}
\label{fig:freq_plot}
\end{figure}

\noindent
\textbf{Base-to-novel generalization:}
To test the generalization ability of various approaches to novel classes, we introduce a \emph{base-to-novel generalization} setting for video action recognition. A dataset $D_S$ with labels $Y_S=\{y_i\}_{i=0}^{k}$ is split into base and novel classes, $Y_B$ and $Y_N$ such that $Y_B \cup Y_N = Y_S$ and $Y_B \cap Y_N = \phi$. The model is learned on base classes and evaluated both on base and novel classes. The proposed base and novel split categorizes the total categories into two equal halves, where the most frequently occurring classes are grouped as the base classes. Fig.~\ref{fig:freq_plot} shows the base-novel splits of Kinetics-400 \cite{k400} and SSv2 \cite{goyal2017ssv2}.

\noindent \textbf{Few-shot setting}: 
We use this setting to test the learning capacity of the model under limited supervision. For a dataset $D_S$ with labels $Y_S=\{y_i\}_{i=0}^{k}$, a general $K$-shot data is created, where K-samples are randomly sampled from each category $y_i \in Y_S$ for training. We use  $K=$ 2, 4, 8 and 16 shots. Validation set of $D_S$ is used for evaluation.

\noindent \textbf{Fully-supervised setting}:
This is the conventional setting for supervised approaches where for a dataset $D_S$ with labels $Y_S=\{y_i\}_{i=0}^{k}$, model is trained on all training examples and evaluated on the respective test set. 

\begin{figure}[!b]
\centering
{\includegraphics[width=0.7\columnwidth]{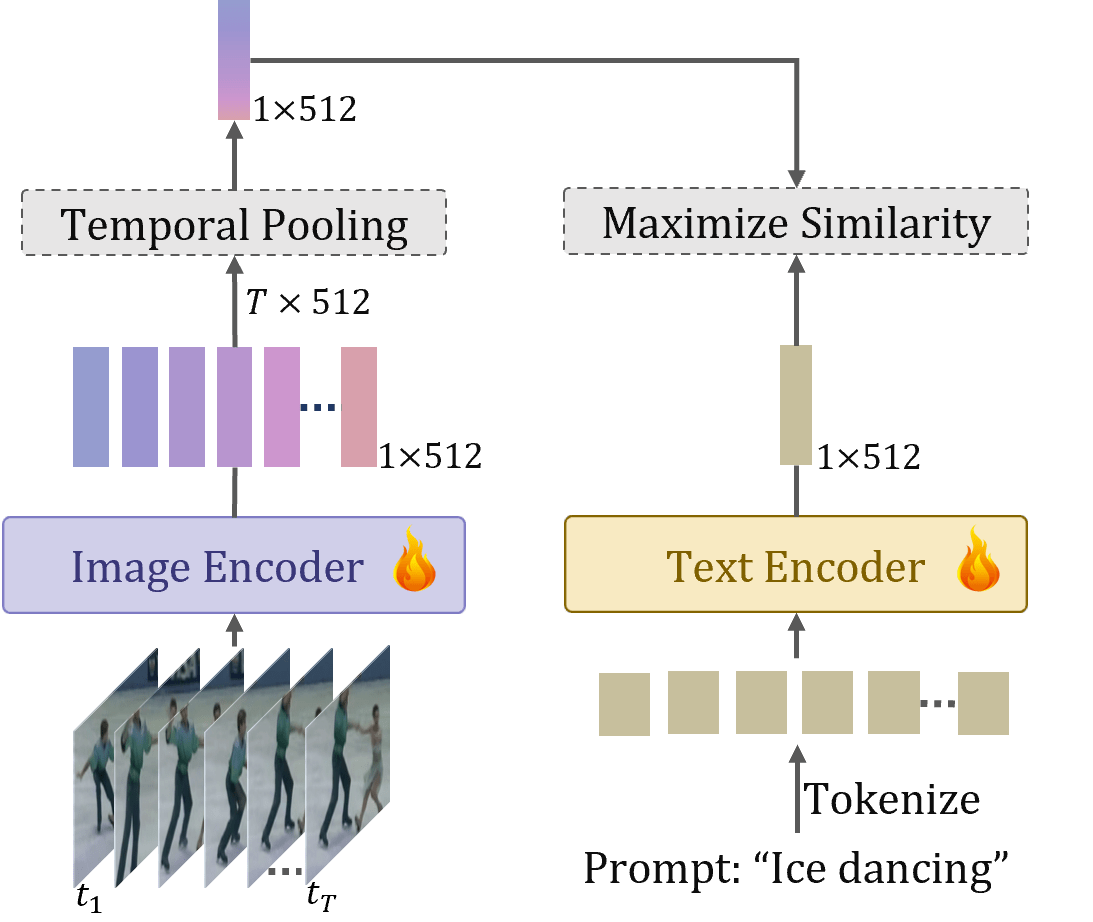}}
\vspace{-0.1em}
\caption{Overview of our simple baseline ViFi-CLIP for adapting CLIP~\cite{clip} to videos. We fine-tune CLIP on videos with minimal design changes that do not include modality specific components which we find to degrade the generalization ability of CLIP~\ref{main_experiments}. Simple frame-level late feature aggregation via temporal pooling allows the exchange of temporal cues in the CLIP representation.}
\label{fig_1:block_giagram}
\vspace{-0.05in}
\end{figure}

\section{Video Finetuned CLIP}
\label{baseline}
\noindent  As training vision-language (VL) models on video-caption pairs is expensive, the availability of large-scale pretrained video VL models is limited. A reliable alternative explored in literature is adaptation of large-scale pretrained image-based VL models, such as CLIP~\cite{clip}, for video downstream tasks. Considering the modality gap, prior methods have explored the use of various specialized attention-based components that instill communication across frames and modules to integrate the information from multiple frames~\cite{wang2021actionclip, ni2022expanding, ju2022prompting}. On the contrary, we explore the capability of a simple baseline called ViFi-CLIP (Video Fine-tuned CLIP) for adapting CLIP~\cite{clip} to video domain. Fig.~\ref{fig_1:block_giagram} illustrates an overview of the proposed baseline ViFi-CLIP.

With the additional temporal information in videos, the important question is how to leverage this information into image-based CLIP model. We explore the capability of \emph{full fine-tuning of CLIP} to bridge the modality gap in video domain. ViFi-CLIP fine-tunes both image and text encoder.

Given a video sample $V_i \in \mathbb{R}^{T \times H \times W  \times C}$ with $T$ frames, and corresponding text label $Y$, the CLIP image encoder encodes the $T$ frames independently as a batch of images and produce frame level embeddings $x_i \in \mathbb{R}^{T \times D}$. These frame-level embeddings are average-pooled to obtain a video-level representation $v_{i} \in \mathbb{R}^{D}$. We refer to this as temporal pooling as this operation implicitly incorporates temporal learning via aggregation of multiple frames.

The CLIP text encoder encodes the class $Y$, wrapped in a prompt template such as ‘\texttt{a photo of a $<$category$>$}’ to produce text embedding $t \in \mathbb{R}^{D}$. 
For a batch of videos, the cosine similarity $sim(.)$, between all the video-level embeddings $v_i$ and the corresponding text embeddings $t_i$ is maximized to fine-tune the CLIP model via cross-entropy (CE) objective with a temperature parameter $\tau$,
\begin{align*}
\Lagr = -\sum_{i} \log\frac{\text{exp}(sim(v_i,t_i)/\tau)}{\sum_{j}\text{exp}(sim(v_i, t_{j})/\tau)}.
\end{align*}

\noindent \textbf{Experimental setup:}
We use ViT-B/16 based CLIP model for our experiments. For zero-shot, base-to-novel and few-shot settings, we use 32 sparsely sampled frames with single view evaluation. In fully supervised setting, we use 16 frames and multi-view inference with 4 spatial crops and 3 temporal views.  We conduct our analysis on five action recognition benchmarks: Kinetics-400 and 600~\cite{k400, k600}, HMDB-51~\cite{kuehne2011hmdb}, UCF-101~\cite{soomro2012ucf101} and Something Something V2 (SSv2)~\cite{goyal2017ssv2}. See Appendix \ref{appendix:iml_details} and \ref{appendix:experiment_settings} for more details.

\begin{SCtable*}[][!ht]
\setlength{\tabcolsep}{5.5pt}
\scalebox{0.84}{

\begin{tabular}{l cccc|cccc|cccc}
  \toprule
  \multirow{2}{*}{Model}  &\multicolumn{4}{c}{HMDB-51} & \multicolumn{4}{c}{UCF-101} & \multicolumn{4}{c}{SSv2}\\
  \cmidrule(lr){2-13}
  & $K$=$2$ & $K$=$4$ & $K$=$8$ & $K$=$16$ & $K$=$2$ & $K$=$4$ & $K$=$8$ & $K$=$16$ & $K$=$2$ & $K$=$4$ & $K$=$8$ & $K$=$16$\\
  \midrule
  \rowcolor{tabhighlight}\multicolumn{13}{c}{Adapting pre-trained image VL models} \\
  \midrule
  Vanilla CLIP~\cite{clip} & 41.9	&41.9	&41.9	&41.9	&63.6	&63.6	&63.6	&63.6	&2.7	&2.7	&2.7	&2.7\\
  ActionCLIP~\cite{wang2021actionclip}  & 47.5	& \underline{57.9}	&57.3	&59.1	&70.6	&71.5	&73.0	& \underline{91.4}	&4.1	& \underline{5.8}	& \underline{8.4}	& \underline{11.1}\\
  XCLIP~\cite{ni2022expanding}        &  \underline{53.0}	&57.3	& \underline{62.8}	& \underline{64.0}	&48.5	&75.6	&83.7	& \underline{91.4}	&3.9	&4.5	&6.8	&10.0\\
  A5~\cite{ju2022prompting} & 39.7	& 50.7	& 56.0	&62.4	& \underline{71.4}	& \underline{79.9}	&  \underline{85.7}	& 89.9	&  \underline{4.4}	& 5.1	& 6.1	& 9.7 \\
  \midrule
 \rowcolor{tabhighlight}\multicolumn{13}{c}{Tuning pre-trained image VL models} \\
  \midrule
  CLIP image-FT & 49.6	&54.9	&57.8	&62.0	&74.4	&79.1	&85.3	&90.5	&4.9	&6.0	&7.2	&10.4 \\
  CLIP text-FT & 54.5	&61.6	&63.1	&65.0	&80.1	&82.8	&85.8	&88.1	&6.2	&6.1	&6.3	&9.1\\
  ViFi-CLIP      & \textbf{57.2}	&\textbf{62.7}	&\textbf{64.5}	&\textbf{66.8}	&\textbf{80.7}	&\textbf{85.1}	&\textbf{90.0}	&\textbf{92.7}	&\textbf{6.2}	&\textbf{7.4}	&\textbf{8.5}	&\textbf{12.4} \\
                  &  \textcolor{MidnightBlue}{{+4.2}} &  \textcolor{MidnightBlue}{{+4.8}}  &  \textcolor{MidnightBlue}{{+1.7}}  &  \textcolor{MidnightBlue}{{+2.8}}  &  \textcolor{MidnightBlue}{{+9.3}}  &  \textcolor{MidnightBlue}{{+5.2}}  &  \textcolor{MidnightBlue}{{+4.3}}  &  \textcolor{MidnightBlue}{{+1.3}}  &  \textcolor{MidnightBlue}{{+1.8}}  &  \textcolor{MidnightBlue}{{+1.6}}  &  \textcolor{MidnightBlue}{{+0.1}}  &  \textcolor{MidnightBlue}{{+1.3}} \\
  \bottomrule                             
\end{tabular}

}
\caption{\textbf{Few-shot setting:} We compare ViFi-CLIP with approaches that explicitly adapt CLIP for video action-recognition on HMDB-51, UCF-101 and SSv2. Gains over the best previous methods that adapt CLIP are indicated in \textcolor{MidnightBlue}{blue} and \underline{underlined} the second best.}
\label{p1_few_shot}
\end{SCtable*}

\subsection{ViFi-CLIP Generalizes Well!}
\label{main_experiments}
\noindent When adapting CLIP to video tasks that demand high generalization ability, two key elements must be satisfied: \romannumeral 1) modality gap should be bridged by adapting image-based CLIP for video domain \romannumeral 2) modality adaptation must happen without hurting the in-build generalization. To analyze the generalization ability of the simple CLIP fine-tuning approach, we evaluate two problem settings:~1) zero-shot setting to evaluate the cross-dataset generalization, and 2) base-to-novel generalization to test performance on novel categories. The later setting  has not been studied before for videos and we introduce new base-to-novel splits for videos. Further details of the splits are given in Appendix~\ref{appendix:datasets}.

\noindent\textbf{{(\romannumeral 1) Zero-shot Setting}}: We investigate the cross-dataset generalization ability of the simple baseline, ViFi-CLIP, in a zero-shot setting. We train the model on a large video action recognition dataset, Kinetics-400 and evaluate across different datasets, HMDB-51, UCF-101 and Kinetics-600. In Table~\ref{p1_zeroshot}, we compare ViFi-CLIP with: 1) uni-modal methods that are specifically designed for zero-shot action recognition, and 2) models that adapt image-based multi-modal VL models for video action recognition. The direct zero-shot evaluation of vanilla CLIP shows impressive generalization performance as compared to uni-modal methods. Further, adapting CLIP with video-specific components helps in improving the generalization in most of the scenarios, indicating the importance of bridging the modality gap. However, the simple fine-tuning approach shows better capability to bridge the domain gap, without disrupting the generalization learned in the pretraining stage of CLIP. Note that we also fine-tune image and text encoders (denoted with CLIP image-FT and CLIP text-FT respectively) and compare with fully fine-tuned CLIP (ViFi-CLIP) where the latter gives stronger generalization due to better alignment of visual and text representations on video tasks. ViFi-CLIP achieves consistent gains of +6.7\%, +4.8\% and +4.5\% in HMDB-51, UCF-101 and K-600 respectively.

\noindent\textbf{{(\romannumeral 2) Base-to-Novel Generalization Setting}}: In Table~\ref{tab2}, we evaluate the generalization from base to novel classes on four datasets, K-400, HMDB-51, UCF-101 and SSv2. In comparison to XCLIP~\cite{ni2022expanding} and ActionCLIP{~\cite{wang2021actionclip}} which use additional components to model video-specific inductive biases, ViFi-CLIP with minimal design modifications provides better base accuracy, and shows noticeable gains in novel accuracy. It provides a better base-to-novel trade-off with an overall best harmonic mean on all datasets. Further, ViFi-CLIP shows better understanding of scene dynamics even on temporally-challenging datasets like SSv2.

\subsection{CLIP directly adapts to Video tasks}
\noindent
We explore the capability of a simple fine-tuning approach in bridging the domain gap on supervised video action recognition tasks under different experimental settings: 1) few-shot learning, 2) fully-supervised setting. 

\begin{table}[b!]
	\centering\setlength{\tabcolsep}{2pt}
	\scalebox{0.87}{
		\begin{tabular}{lcccccc}
			\toprule
			Method & Frames & Top-1 &  Top-5 & Views & GFLOPs & TP \\ \midrule
			\rowcolor{tabhighlight} \multicolumn{7}{c}{{Uni-modal architectures}} \\
			\midrule
			Uniformer-B~\cite{li2022uniformer}   &   32   &   83.0 & 95.4 &  4 $\times$ 3  & 259 & - \\
			TimeSformer-L~\cite{timesformer2021} &   96   &   80.7 & 94.7 &  1 $\times$ 3 & 2380 & -  \\
			Mformer-HR~\cite{patrick2021keeping} &   16   &   81.1 & 95.2 &  10 $\times$  3  &  959 & - \\
			Swin-L~\cite{liu2021video}    		 &   32   &   83.1 & 95.9 &  4 $\times$ 3  & 604 & - \\
			\midrule
			\rowcolor{tabhighlight} \multicolumn{7}{c}{{Adapting pre-trained image VL models}} \\
			\midrule
		    ActionCLIP~\cite{wang2021actionclip} & 32 &   83.8  & 96.2 &  10 $\times$ 3   & 563 & 67.7\\
		    X-CLIP~\cite{ni2022expanding} & 16 & \textbf{84.7} &  \textbf{96.8} & 4 $\times$ 3 & 287 & 58.5  \\
			A6~\cite{ju2022prompting}  & 16 &  76.9  & 93.5  &  - & - & -  \\
			\midrule
			\rowcolor{tabhighlight} \multicolumn{7}{c}{{Tuning pre-trained image VL models}} \\
			\midrule
		    CLIP image-FT &  16  &   82.8  & 96.2 & 4 $\times$ 3 & 281 & 71.1\\
			CLIP text-FT	&  16  &   73.1  & 91.2 & 4 $\times$ 3 & 281 & 71.1 \\
			ViFi-CLIP   	&  16  &   \underline{83.9}  &  \underline{96.3} &  4 $\times$ 3 & 281 & 71.1 \\ 
			\bottomrule
		\end{tabular}}
	\caption{\textbf{Fully-supervised setting:} We compare ViFi-CLIP with uni-modal methods and models specifically designed to adapt CLIP for video tasks on Kinetics-400. In addition to accuracy, we report FLOPs and throughput (TP).
	\label{p1_fullysupervised}
	}
\end{table}

\begin{figure*}[!ht]
\centering
{\includegraphics[width=0.97\textwidth]{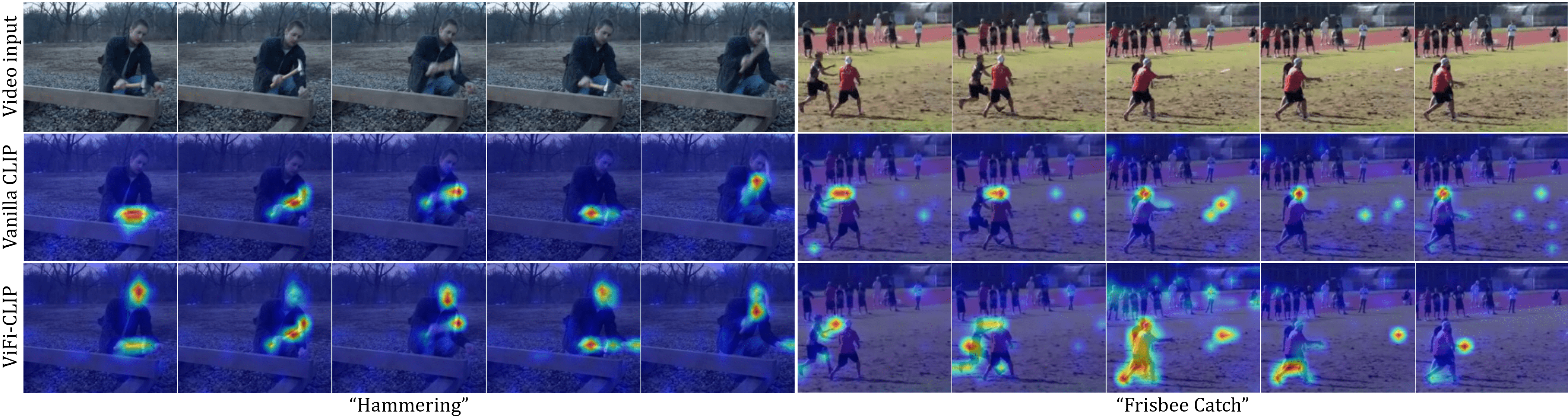}}
\caption{\textbf{Attention map visualizations} of  ViFi-CLIP in comparison with vanilla CLIP on two examples from UCF-101 validation set. ViFi-CLIP learns inter-object relationships and scene-dynamics from temporal cues and focuses on fast-moving parts and objects, thereby demonstrating the ability to encode video specific information. \textbf{(Left):} An example on action class `hammering'. While vanilla CLIP focuses only on the object (hammer), ViFi-CLIP attends to the interaction between the person and the object. \textbf{(Right):} Example on `frisbee catch' category. Vanilla CLIP uses only appearance cues and confuses a hat with the frisbee, while ViFi-CLIP focuses on the players and pays more attention on fast-moving parts like the hands and legs of the players, and correctly locates the frisbee. }
\label{fig_3:attention_vis}
\end{figure*}

\noindent\textbf{{(\romannumeral 1) Few-shot Setting}}: In Table~\ref{p1_few_shot}, we show the effect of ViFi-CLIP in the few-shot setting along with methods that adapt CLIP for videos. We note that ViFi-CLIP consistently improves performance with increasing shots. Across the three datasets HMDB-51, UCF-101 and SSv2 in all the shots ($K=$ 2, 4, 6, 8), it provides better performance against all the compared methods. Interestingly, it achieves relatively larger gains in extremely limited data scenarios demonstrating robustness towards overfitting. For example, it achieves gains of +9.3\% and 4.2\% in UCF-101 and HMDB-51 respectively compared to prior best methods.

\noindent\textbf{{(\romannumeral 2) Fully-supervised Setting}}: We compare the performance of ViFi-CLIP trained on Kinetics-400 with uni-modal video-specific models and other methods that tailor CLIP for videos in Table~\ref{p1_fullysupervised}. The simple approach of \textit{fully fine-tuning} CLIP provides competitive performance in comparison to methods that use additional carefully designed learnable components for video-specific temporal modeling. Further, the ablation of fine-tuning image-encoder and text-encoder indicates the effectiveness of fine-tuning the full CLIP model to address the domain gap.

\subsection{How simple fine-tuning bridges domain gap?}
\noindent Having shown the effectiveness of ViFi-CLIP for adapting  CLIP for video action recognition, we explore how this approach encodes video specific information that enables bridging the modality gap. We conduct our experiments by ablating on the fusion mechanism that is used to combine frame-level information. In the proposed baseline, we adopt an embedding-level fusion, where individual frames are encoded by the image encoder, and the resulting image embeddings are then fused together to obtain a video-level visual representation. We explore two alternate fusion mechanisms; 1) \emph{Decision-level fusion:} image embeddings from individual frames are used separately to compute similarity (logits) with the corresponding text embeddings. The frame-level logits are then averaged to obtain video-level logits. 2) \emph{Image-level fusion:} the frames of a video are considered as individual images, and losses are computed across each frame, thus removing all temporal information.

\begin{table}[!ht]
\setlength{\tabcolsep}{1mm}{
\resizebox{0.99\columnwidth}{!}{
\begin{tabular}{lcccccc}
\toprule
\rowcolor{tabhighlight} Method & K400-tiny & SSv2 & HMDB-51 & UCF-101\\
\midrule
Vanilla CLIP~\cite{clip}     & 51.6	& 2.7 &	41.9 &	63.6	\\
Decision-level fusion  &  {74.6}&	10.8 &	62.2 &	90.5 \\
Image-level fusion & 74.0&	{11.2} &	 {64.4} &	 {91.7} \\
Embedding fusion &  \textbf{76.9} &	\textbf{12.4} &	\textbf{66.8} &	\textbf{92.7}\\
\bottomrule
\end{tabular}
}}
\caption{Analysis on different fusion mechanisms that integrate temporal information in the model. ViFi-CLIP uses an embedding-level fusion, where representations of multiple frames are combined together to integrate the temporal information.}
\label{table5:ablation_temporal}
\end{table}

In Table~\ref{table5:ablation_temporal}, we compare the two fusion mechanisms with our simple embedding level fusion across four datasets, Kinetics-400 tiny (a smaller split of full K-400), few-shot ($K=16$) splits of SSv2, HMDB-51 and UCF-101. The analysis shows that fusing the frame-level embeddings helps the model learn the temporal relations between different frames, thus implicitly establishing inter-frame communication. We note a significant gain on SSv2 with embedding-level fusion that further supports the intuition, as SSv2 demands rich temporal modeling due to its fine-grained actions as compared to Kinetics-400. 

\subsection{How effective are the video-specific representations learned during simple fine-tuning?}
\noindent We conduct qualitative analysis on the generalization performance of ViFi-CLIP for the zero-shot setting, as shown in Fig.~\ref{fig_1:tsne_plot_into}. The t-SNE visualisation of video-embeddings from ViFi-CLIP are compared with vanilla CLIP, other alternatives (CLIP image-FT and CLIP text-FT) and prior state-of-the art method XCLIP~\cite{ni2022expanding}. The feature representations improve with the fine-tuning of either text or image-encoder over vanilla CLIP. When both text and image encoder are tuned, the learned representations further improve and show better separability in the latent space. Additionally, we note that ViFi-CLIP achieves competitive performance when compared with XCLIP, that requires dedicated components to model temporal information.

\begin{figure*}[!ht]
\centering
{\includegraphics[width=0.99\textwidth]{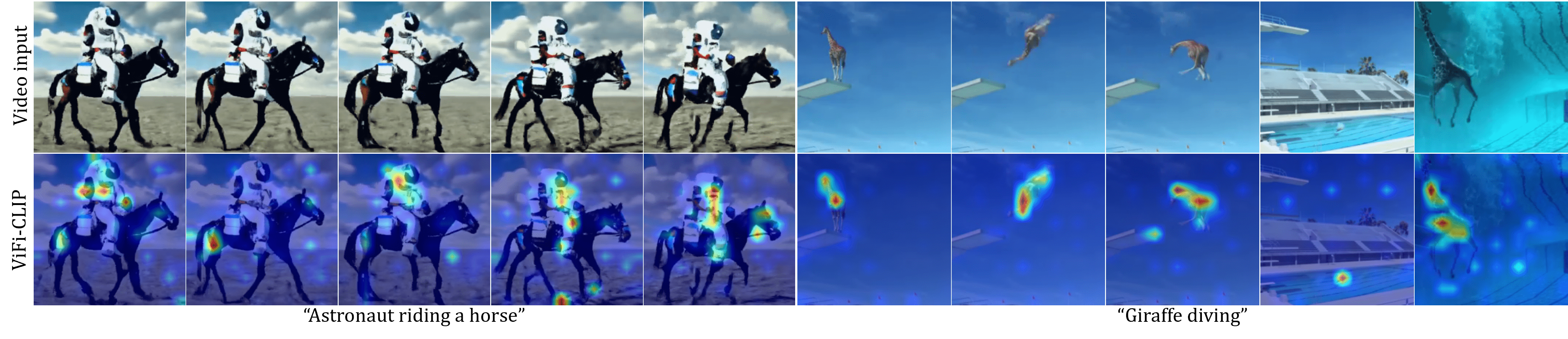}}
\caption{\textbf{Generalization to out-of-distribution examples}: Attention map visualizations from ViFi-CLIP shows impressive generalization. \textbf{(Left):} Visualization on a diffusion model generated (synthetic) video from Imagen~\cite{saharia2022photorealistic} shows how ViFi-CLIP focuses on the body of the astronaut, the horse and its moving feet. \textbf{(Right):} Example of a rare scenario `\txt{giraffe diving}'. ViFi-CLIP consistently attends to the giraffe at various locations: diving board, air, water-surface and under the pool.}
\label{fig_5:attention_vis_extreme}
\end{figure*}

To better understand what temporal relations are captured by fine-tuning CLIP on a video-dataset, we present attention map visualizations in Fig.~\ref{fig_3:attention_vis}. Our empirical analysis stipulates two interesting findings: 1) ViFi-CLIP learns inter-object relationships from temporal cues to recognize the action. For example, in Fig.~\ref{fig_3:attention_vis} (left), the model focuses not only on the object (hammer), but attends to the interaction between the person and the object. 2) ViFi-CLIP focuses on scene dynamics and moving objects. In Fig.~\ref{fig_3:attention_vis} (right), ViFi-CLIP attends to the fast-moving parts of the scene and learns to attend to salient parts of the body important for temporal understanding such as the legs and hands of the players. Intuitively, these observations indicate that temporal relations can be implicitly modeled by simply fine-tuning CLIP on a video-dataset. Additionally, we test ViFi-CLIP on extreme out-of-distribution examples. Attention maps shown in Fig.~\ref{fig_5:attention_vis_extreme} demonstrate good generalization. It supports the claim that ViFi-CLIP learns temporal relations and aids in boosting the generalization of CLIP~\cite{clip} by keeping CLIP image and text encoders intact.
\vspace{-0.1in}
\subsection{Is fine-tuning efficient w.r.t adapting CLIP?}
\noindent Being a competitive alternative to other methods in-terms of accuracy, we further study the compute complexity of ViFi-CLIP in comparison to other methods.

\begin{table}[!ht]
\centering
\setlength{\tabcolsep}{3.55mm}{
\resizebox{0.9\columnwidth}{!}{
\begin{tabular}{lcccc}
\toprule
\rowcolor{tabhighlight} Method & GFLOPs & TP & Params (M)\\
\midrule
ActionCLIP~\cite{wang2021actionclip} &  563 &  67.7  & 168.5\\ 
XCLIP~\cite{ni2022expanding}  & 287	& 58.5 &	 131.5\\
ViFi-CLIP & 281 & 71.1 & 124.7\\
\bottomrule
\end{tabular}
}}
\caption{Compute comparison of ViFi-CLIP with methods that adapt CLIP with additional components. Throughput per view (TP) is measured using a single A100 GPU. ViFi-CLIP enjoys efficiency in-terms of GFLOPs, throughput and parameter count.}
\label{vifi_complexity}
\vspace{-0.1in}
\end{table}

\noindent Table~\ref{vifi_complexity} shows that ViFi-CLIP provides high throughput (TP) of 71.1 images/sec as compared to other methods that adapt CLIP for videos. This is attributed mainly due to its simple design that avoids using any additional video-specific components. This also leads to lower FLOPs and fewer training parameters as compared to other approaches.

\section{Bridge and Prompt in low-data regimes}
\label{prompting}
\noindent ViFi-CLIP shows that a simple fine-tuning approach is effective in bridging the domain gap in video. However, fine-tuning the CLIP model may not always be feasible as it requires training large number of parameters. Particularly in case of low-data regimes, where availability of training data is extremely limited, we explore an important question: How can one efficiently steer CLIP towards various downstream tasks, after bridging the modality gap, that favours both effectiveness and efficiency in-terms of performance and compute respectively? We explore a two-stage framework, `\emph{bridge and prompt}': \romannumeral 1) The model is first fine-tuned on a video dataset to bridge the modality gap, \romannumeral 2) Model is adapted to downstream tasks for better generalization through context optimization via prompting. Here the entire model is frozen, and prompts are adapted and learned for a specific task. Ju \etal propose a strong baseline that learns task-specific vision prompts for adapting CLIP for video tasks~\cite{ju2022prompting} and use lightweight transformers for temporal modeling. Although this efficient prompting technique proves to adapt CLIP for video tasks, the model struggles to generalize towards unseen classes due to the late fusion through the transformer layers. 

\begin{figure}[!b]
\centering
{\includegraphics[width=0.73\columnwidth]{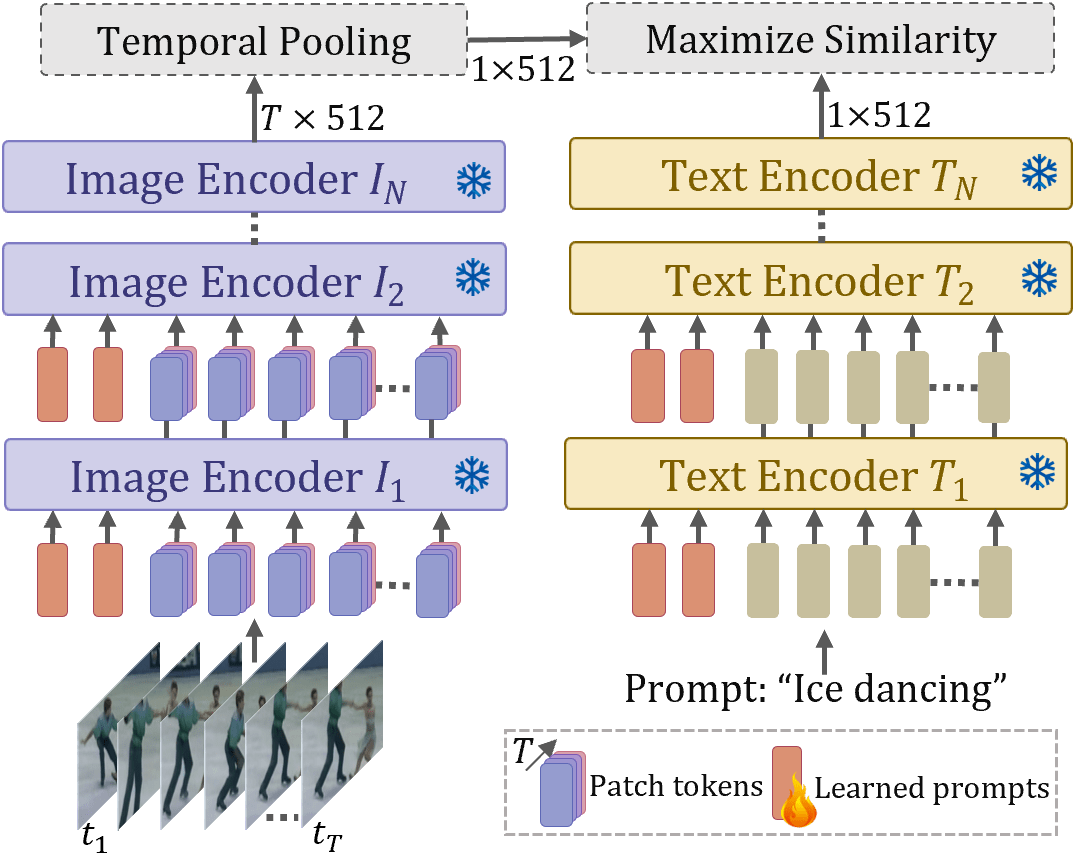}}
\caption{We use Vision-Language prompting approach to adapt CLIP for videos. Vision and textual prompt tokens are used in the vision and language branch of CLIP which are the model's only learnable parameters. These prompts steer CLIP towards downstream tasks in low data regime scenarios without losing the original generalization ability of CLIP. Deep contextual prompts are used in both branches across multiple transformer layers.}
\label{fig_2:prompt_block_giagram}
\end{figure}

\begin{SCtable*}[][!ht]
\setlength{\tabcolsep}{6pt}
\scalebox{0.82}{
\begin{tabular}{l cccc|cccc|cccc}
  \toprule
  \multirow{2}{*}{Model}  &\multicolumn{4}{c}{HMDB-51} & \multicolumn{4}{c}{UCF-101} & \multicolumn{4}{c}{SSv2}\\
 \cmidrule(lr){2-13}
  & $K$=$2$ & $K$=$4$ & $K$=$8$ & $K$=$16$ & $K$=$2$ & $K$=$4$ & $K$=$8$ & $K$=$16$ & $K$=$2$ & $K$=$4$ & $K$=$8$ & $K$=$16$\\
  \midrule
  Vanilla CLIP~\cite{clip} & 41.9	&41.9	&41.9	&41.9	&63.6	&63.6	&63.6	&63.6	&2.7	&2.7	&2.7	&2.7\\
  ActionCLIP~\cite{wang2021actionclip}  & 54.3	&56.2	&59.3	&66.1	&76.7	&80.4	&87.6	&91.8	&4.8	&6.9	&9.1	&12.3 \\
  XCLIP~\cite{ni2022expanding} &  \underline{60.5}&	 \textbf{66.8}&	 \underline{69.3}&	 \underline{71.7}&	 \underline{89.0}&	 \underline{91.4}&	 \underline{94.7}&	 \underline{96.3}&	 \underline{6.6}	&  \underline{7.8}&	 \underline{9.9}	&  \textbf{13.7}\\
  A5~\cite{ju2022prompting} & 46.7&	50.4&	61.3&	65.8&	76.3&	84.4&	90.7&	93.0&	4.5&	  6.7&	7.2	& 9.5\\
  \midrule
  VL Prompting  &\textbf{63.0}	& \underline{65.1} &	\textbf{69.6} &	\textbf{72.0} &	\textbf{91.0} &	\textbf{93.7} &	\textbf{95.0} &	\textbf{96.4} &	\textbf{6.7}	& \textbf{7.9} &	\textbf{10.2} &	\underline{13.5}\\
  \bottomrule                             
\end{tabular}
}
\caption{VL prompting effectively improves over other methods in few-shot setting. Models are pretrained on Kinetics-400 to bridge the modality gap.
}
\label{prompt_few_shot}
\end{SCtable*}
\begin{SCtable*}[][!ht]
\vspace{-0.2in}
\setlength{\tabcolsep}{6pt}
\scalebox{0.87}{
\begin{tabular}[t]{l ccc|ccc|ccc}
\toprule
&      \multicolumn{3}{c}{HMDB-51}      &\multicolumn{3}{c}{UCF-101}    & \multicolumn{3}{c}{SSv2} \\
\cmidrule(lr){2-10}
Method  & Base & Novel & HM                & Base & Novel & HM         & Base & Novel & HM   \\ \midrule
Vanilla CLIP~\cite{clip}      & 53.3 & 46.8 & 49.8           & 78.5 &63.6 & 70.3      & 4.9 & 5.3 & 5.1 \\ 
ActionCLIP~\cite{wang2021actionclip}        & 69.0 &  \textbf{57.2} &  \underline{62.6}        & 85.6 &  \textbf{75.3} & 80.1           & 8.1 & 8.7 & 8.4 \\
XCLIP~\cite{ni2022expanding}              &  \underline{75.8} & 52.0 & 61.7             & 95.4 & 74.0 &  \underline{83.4}       &  \underline{14.2} &  \underline{11.0} &  \underline{12.4} \\
A5~\cite{ju2022prompting}     & 70.4 & 51.7 & 59.6            & \underline{95.8} & 71.0 & 81.6       & 12.9 & 5.7 & 7.9\\ \midrule
VL prompting     & \textbf{77.1}	& \underline{54.9} & \textbf{64.1} & \textbf{95.9} &	\underline{74.1} &	\textbf{83.6} & \textbf{15.8}	& \textbf{11.5} &	\textbf{13.3} \\ \bottomrule
\end{tabular}
}
\caption{
Comparison of VL prompting approach in base-to-novel generalization setting. VL prompting shows consistent gains on base classes while also improving on novel classes. It performs competitive even against fine-tuning based approaches~\cite{wang2021actionclip,ni2022expanding}. All models are first pretrained on Kinetics-400.
}
\label{prompt_base_to_novel}

\end{SCtable*}

To this end, we develop an extended baseline that efficiently adapts CLIP in low-data regimes via vision-language (VL) prompt learning for videos. Fig.~\ref{fig_2:prompt_block_giagram} shows the overall architecture of our proposed framework. In contrast to previous approaches that learn prompts only at the language branch for video adaption~\cite{ju2022prompting}, we use a vision-language prompt learning design where prompt vectors are learnt on both vision and language branch. Moreover, we introduce prompts at deeper layers of both encoders, to capture hierarchical contextual representations. 

\subsection{Prompting is effective on fine-tuned CLIP}
\noindent We compare our VL prompting with other methods that adapt CLIP for videos.
Following the first stage of our `bridge and prompt' approach, all methods are pretrained on Kinetics-400 and then evaluated in two problem settings: few-shot transfer and base-to-novel generalization. Note that ActionCLIP~\cite{wang2021actionclip} and XCLIP~\cite{ni2022expanding} fine-tune CLIP encoders on the corresponding datasets.

\noindent\textbf{(\romannumeral 1) Few-shot Setting}: Table~\ref{prompt_few_shot} shows the results for few-shot transfer. Vanilla CLIP~\cite{clip} is the lower bound and A5~\cite{ju2022prompting} is most similar to our approach as it only adapts prompts and few transformer layers, keeping the CLIP model frozen. We note that VL prompting consistently provides better performance over A5 and even performs competitively against fine-tuning approaches. Particularly for extreme cases where $K=2$, it provides the best results by providing absolute gains of 2.5\% and 2\% over XCLIP on HMDB-51 and UCF-101 respectively. This suggests the significance of learning prompts in low-data regimes.

\noindent\textbf{(\romannumeral 2) Base-to-Novel Generalization Setting}: In Table~\ref{prompt_base_to_novel}, we compare the results on base-to-novel setting among different methods. In comparison to vanilla CLIP, all fine-tuning approaches improves their generalization ability for novel classes. We note that vision-language (VL) prompting provides competitive performance against prior prompting~\cite{ju2022prompting} designs and highly competitive fine-tuning methods~\cite{ni2022expanding, wang2021actionclip} without any video-specific attention modules. The results suggest that VL prompting is effective in steering pretrained CLIP model towards downstream tasks without compromising on generalization. 

\subsection{Is prompting efficient w.r.t CLIP adaptation?}
\noindent We perform an analysis on complexity of various methods in the low-data regime as detailed in Table~\ref{prompting_complexity}. CLIP adaptation methods provide less throughput due to the use of video-specific learnable components in addition to the vanilla CLIP model. A5~\cite{ju2022prompting} requires fewer FLOPs but achieves less throughput, due to the additional transformer blocks for temporal modeling. On the other hand, VL prompting shows higher efficiency in terms of throughput and comparable FLOPs as compared to prior approaches.

\begin{table}[!ht]
\centering
\setlength{\tabcolsep}{6mm}{
\resizebox{0.80\columnwidth}{!}{
\begin{tabular}{lccc}
\toprule
\rowcolor{tabhighlight} Method & GFLOPs & TP\\
\midrule
ActionCLIP~\cite{wang2021actionclip} &  563 &  67.8 \\ 
XCLIP~\cite{ni2022expanding}  & 287	& 58.5 \\
A5~\cite{ju2022prompting} & 284 & 62.5 \\
VL prompting & 287 & 71.6\\
\bottomrule
\end{tabular}
}}
\caption{Computational complexity comparison of VL prompting with other methods in-terms of GFLOPs and throughput (TP).}
\label{prompting_complexity}
\vspace{-0.2in}
\end{table}

\section{Conclusion}
\noindent This work shows the significance of an often neglected but simple baseline for transferring image-based CLIP model to video domain. We demonstrate that simply fine-tuning both the vision and text encoders on video data performs favourably on supervised as well as generalization tasks.  The results show the scalabiltiy and advantage of a simple solution with respect to sophisticated approaches developed dedicatedly for videos in majority of the settings. In cases where fine-tuning is not possible, we also propose a bridge and prompt scheme that uses the video fine-tuned representations to quickly adapt to downstream video applications. 

{\small
\bibliographystyle{ieee_fullname}
\bibliography{egbib}
}
\newpage
\newpage
\appendix

\begin{center}
\textbf{\Large Supplemental Material}
\end{center}

\noindent We provide supplementary material that provides additional details and further qualitative analysis for the main paper. The contents follow the following order:
\begin{itemize}
    \item Other video-level tasks  (Appendix~\ref{appendix:video_retrieval})
    \item Additional implementation details (Appendix~\ref{appendix:iml_details})
    \item Datasets (Appendix~\ref{appendix:datasets})
    \item Evaluation protocols (Appendix~\ref{appendix:experiment_settings})
    \item Additional Qualitative results (Appendix~\ref{appendix:qualitative_results})
\end{itemize}

\section{Other video-level tasks}
\label{appendix:video_retrieval}
\noindent We scale and evaluate our approach on video retrieval task on MSRVTT (9K) dataset with consistent settings (16 frames using ViT-B/16) as A5 model [17] and show improved performance in Table \ref{video_retrieval}. 

\begin{table}[!ht]
\centering
\setlength{\tabcolsep}{6mm}{
\resizebox{0.80\columnwidth}{!}{
\begin{tabular}{lcc}
        \toprule
        \rowcolor{Gray} Method & \multicolumn{1}{c}{R@1} & \multicolumn{1}{c}{R@5}\\
         \midrule
        Frozen & 31.0 & 59.5 \\
        A5 [17] & 36.7 & 64.6\\
        \small ViFi-Clip & \textbf{44.8} & \textbf{72.4}\\
        \bottomrule
        \end{tabular}
}}
\caption{Comparison of ViFiCLIP with methods that explicitly adapt CLIP for videos on the Video-retrieval task.}
\label{video_retrieval}
\vspace{-0.2in}
\end{table}

\section{Implementation Details}
\label{appendix:iml_details}
\noindent In all our experiments on ViFi-CLIP, and its variants, individually tuned CLIP text encoder (CLIP text-FT) and image encoder (CLIP image-FT), all randomly sampled frames are pre-processed to a spatial size of 224×224. In our experiments, we use handcrafted text prompts with a template ‘\texttt{a photo of a $<$category$>$}’. Following CLIP~\cite{clip}, the maximum number of text tokens is set to 77. We use an AdamW optimizer and weight decay of 0.001. We modify the epochs, batch size and learning rate across the different experimental settings, which are detailed below. 

We conduct our analysis under four experimental settings: zero-shot, base-to-novel generalization, few-shot and fully-supervised. In the \textit{zero-shot setting}, ViFi-CLIP and its variants are trained for 10 epochs on Kinetics-400 dataset with a batch size of 256, and a learning rate of 8e-6. In the \textit{base-to-novel generalization} and the \textit{few-shot} setting, ViFi-CLIP is trained in a few-shot manner, with a batch size of 64, and a learning rate of 2e-6. For the \textit{fully-supervised setting}, we train ViFi-CLIP on the kinetics-400 dataset for 30 epochs with a batch size of 512 and a learning rate of 22e-6.

We implement other baseline methods including A6~\cite{ju2022prompting}, ActionCLIP~\cite{wang2021actionclip} and XCLIP~\cite{ni2022expanding} using their default optimal hyper-parameters as reported in their work. For Efficient prompting~\cite{ju2022prompting}, we use their best performing A6 model for the fully-supervised setting, and use their A5 model in the zero-shot, and base-to-novel generalization and few-shot settings. In case of ActionCLIP~\cite{wang2021actionclip}, we use their best-performing variant \textit{Transf} \cite{wang2021actionclip} in all our experiments.

\section{Dataset details}
\label{appendix:datasets}

\noindent  We conduct our analysis on five established action recognition benchmarks: Kinetics-400~\cite{k400} and Kinetics-600~\cite{k600}, HMDB-51~\cite{kuehne2011hmdb}, UCF-101~\cite{soomro2012ucf101} and Something-Something v2 (SSv2)~\cite{goyal2017ssv2}. 

\vspace{0.1in}
\noindent \textbf{Kinetics-400 and Kinetics-600}: The K-400 datasets contains 400 human action classes comprising video clips taken from various YouTube videos that lasts for about 10 seconds. It contains around 240K training and 20K validation videos. The K-600 is an extension of of K-400, with around 650K video clips covering 600 action categories, consisting of around 410K training and 29K validation videos. 

\vspace{0.1in}
\noindent \textbf{HMDB-51}: The HMDB-51 dataset contains 71K realistic videos collected from different sources spanning 51 action categories. The standard split consist of 3570 training samples and 1530 validation samples. The training and validation are further split into three individual splits, each containing 70 and 30 clips of all action categories for training and validation, respectively.

\vspace{0.1in}
\noindent \textbf{UCF-101}: UCF-101 contains 13K realistic videos collected from YouTube covering 101 action categories that includes five types of action: human-object interaction, body-motion, human-human interaction, playing instrumental music and sports. The standard split trains on 9537 videos and evaluates on 3783 videos, which are grouped into three splits.

\vspace{0.1in}
\noindent \textbf{Something-Something v2 (SSv2)}: The SSv2 dataset is a large collection of video clips of humans performing actions with everyday objects, spanning 174 action categories. The dataset evaluates the capacity of the model on fine-grained actions such as covering something with something or uncovering something, making the dataset more temporally biased as opposed to other datasets. The standard split consist of 168,913 training videos and 24,777 validation videos. We report the top-1 accuracy over the validation split. 

\begin{figure*}[!ht]
\centering
{\includegraphics[width=0.99\textwidth]{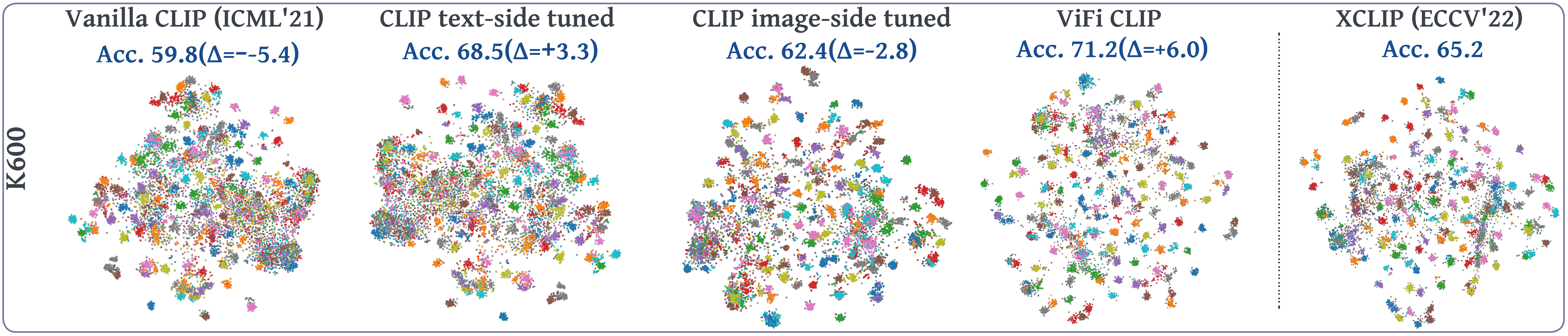}}
\caption{\textbf{t-SNE visualizations for Kinetics-600.} For K600 \cite{k600}, we show the t-SNE visualizations for 160 classes that are non-overlapping with Kinetics-400. The fine-tuned models are trained on Kinetics-400 and evaluated on the non-overlapping classes of Kinetics-600.}
\label{appendix:tsne_visualizations}
\end{figure*}

\section{Evaluation Protocols}
\label{appendix:experiment_settings}
\noindent We conduct our analysis on four different experimental settings:  \emph{zero-shot}, \emph{base-to-novel generalization}, \emph{few-shot} and \emph{fully-supervised} setting. Across these settings, we use a sparse sampling strategy~\cite{wang2016tsn} to sample frames and set the number of frames to 16 or 32 (specified under each setting). Each sampled frame is spatially scaled on the shorter side to 256, with a center crop of 224.

\vspace{0.1in}
\noindent \textbf{Zero-shot setting}: Under the zero-shot setting, models trained on Kinetics-400 are evaluated on three cross datasets, HMBD-51, UCF-101 and Kinetic-600. For HMBD-51 and UCF-101, the methods are evaluated on their corresponding three validation splits and we report the top-1 average accuracy over them. In case of Kinetics-600, we follow ~\cite{chen2021elaborative} and evaluate the methods on 220 categories that are non-overlapping with Kinetics-400. We report top-1 and top-5 average accuracy over three randomly sampled splits, each containing 160 categories. In this setting, we use a single-view inference with 32 frames. 

\begin{figure*}[ht]
\centering
{\includegraphics[width=0.99\textwidth]{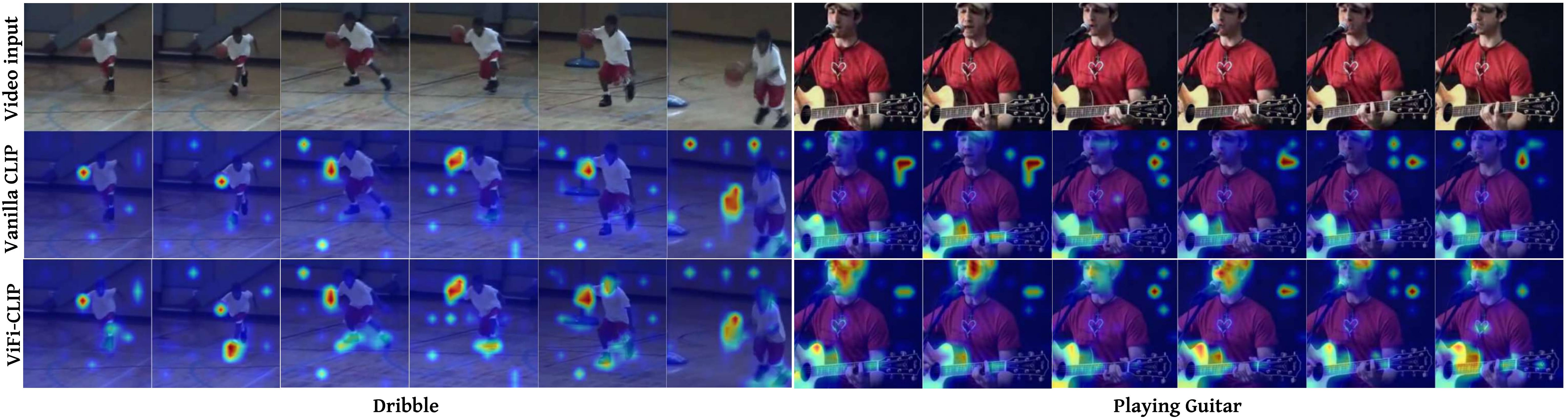}}
\caption{\textbf{Attention map visualizations} of ViFi-CLIP in comparison with vanilla CLIP on two examples from HMDB-51 (left) and UCF-101 (right) validation set. Fine-tuning CLIP on video-datasets in ViFi-CLIP helps the model learn inter-object relationships and scene-dynamics from temporal cues. The model focus on moving objects and fast-moving parts which indicates that ability of ViFi-CLIP to encode video specific information. \textbf{(Left):} An example on an action class with fast motion, `dribble'. While vanilla CLIP focuses only on the ball, ViFi-CLIP attends to the interaction between the player and the object. Moreover, it always focuses on fast-moving parts of the player (legs), thus shows ability to focus on temporal cues. \textbf{(Right):} Example from `playing guitar' category. While vanilla CLIP uses only appearance cues and attends to the guitar, ViFi-CLIP focuses on the interaction between the singer and the guitar, and pays attention on moving parts like lips of the players.}
\label{fig_sup1:sup_attention_vis}
\end{figure*}

\begin{figure*}[ht]
\centering
{\includegraphics[width=0.99\textwidth]{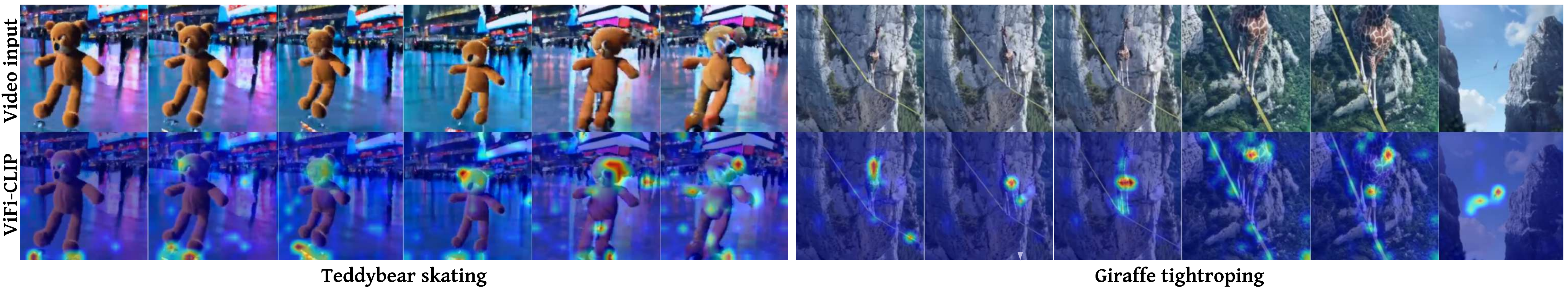}}
\caption{\textbf{Generalization to out-of-distribution examples}: Attention map visualizations from ViFi-CLIP shows good generalization. \textbf{(Left):} Visualization on a synthetically generated video from Imagen~\cite{saharia2022photorealistic} shows how ViFi-CLIP focuses on inter-object relationships, like the teddy-bear and the skating shoes. \textbf{(Right):} Example of a rare scenario `\txt{giraffe tight-roping}'. ViFi-CLIP attends to the giraffe at difference scene variations and additionally focuses on the tight-rope, showing ability to capture inter-object relationships.}
\label{fig_sup2:sup_attention_vis_extreme}
\end{figure*}

\vspace{0.1in}
\noindent \textbf{Base-to-novel setting}: For extensive analysis on the generalization ability of various approaches, we introduce a \emph{base-to-novel generalization} setting for video action recognition tasks, where a model is first trained on a set of \textit{base} (seen) classes in a few-shot manner and evaluated on a set of \textit{novel} (unseen) classes. We present comprehensive generalization analysis on four datasets, Kinetics-400, HMBD-51, UCF-101 and SSv2. For each dataset, we create three training splits, each containing randomly sampled 16-shots of every action category. The split categorizes the total categories into two equal halves, where the the most frequently occurring classes are considered as the base classes, and the rarely occurring categories are taken as the novel classes. Figure~\ref{fig:freq_plot} shows the frequency distribution of the Kinetics-400 and depicts the resulting base-novel split. The model is evaluated on the corresponding validation splits. In case of HMDB-51 and UCF-101, the training and validation considers only their first split, while for Kinetics and SSv2, the models are evaluated on their full validation split. The setting uses 32 frames and follows a single-view inference.

\vspace{0.05in}
\noindent \textbf{Few-shot setting}: The few-shot setting creates a general K-shot split, where K-samples are randomly sampled from each category for training. Specifically, we use 2, 4, 8 and 16 shots for three datasets, HMBD-51, UCF-101 and SSv2. The models are evaluated on the first validation split for HMDB-51 and UCF-101 and the full validation split in case of SSv2. In this setting, we use 32 sparsely sampled frames and evaluate with single-view inference. 

\vspace{0.1in}
\noindent \textbf{Fully-supervised setting}: In the fully-supervised setting, the methods are trained on Kinetics-400 are evaluated on its complete validation set. We use 16 frames and report multi-view inference with three different spatial crops and four temporal clips.

\section{Additional qualitative results}
\label{appendix:qualitative_results}

\noindent
The t-SNE visualizations of video-embeddings in Fig.~1 are computed for the UCF101 (1st col.) and HMDB51 (2nd col.), whereas for K-600 in Fig.~7. Here, each color represents a category. We observe the embeddings of ViFi-CLIP within a category are better separable from others, indicating the effectiveness of the proposed approach to learn suitable video-specific inductive biases. We also evaluate the quality of the clusters in the visualizations. The homogeneity (H), completeness (C), and V-measure (V) computed for vanilla CLIP, XCLIP and ViFi-CLIIP for HMDB51 in table \ref{metrics_tsne}, show trends consistent to our t-SNE visualizations. 

\begin{table}[!ht]
\centering
\setlength{\tabcolsep}{6mm}{
\resizebox{0.80\columnwidth}{!}{
\begin{tabular}{lccc}
\toprule
\rowcolor{Gray} Method & H ($\uparrow$) & C ($\uparrow$) & V ($\uparrow$)\\
\midrule
Vanilla CLIP & 0.84 & 0.86 & 0.85 \\
XCLIP & 0.92 & 0.93 & 0.93 \\
ViFi-CLIP & \textbf{0.94} & \textbf{0.95} &\textbf{ 0.95}\\
\bottomrule
\end{tabular}
}}
\caption{Metrics evaluating the quality of clusters in the t-SNE visualizations.}
\label{metrics_tsne}
\vspace{-0.2in}
\end{table}

In Fig.~\ref{appendix:tsne_visualizations}, we show additional t-SNE visualizations for the zero-shot evaluation of Kinetics-600~\cite{k600}. We compare video embedding of vanilla CLIP and its fine-tuned variants with XCLIP~\cite{ni2022expanding}. The models are trained on Kinetics-400~\cite{k400} and then evaluated directly on non-overlapping classes of K600. For ViFi-CLIP, the video embeddings of different classes show better separability among all other approaches. ViFi-CLIP finetunes both the text and vision encoder of CLIP, achieves better generalization performance and provides a gain of +6\% on Kinetic-600 in comparison to the recent state of the art method XCLIP\cite{ni2022expanding}.

To analyze the type of temporal information captured by ViFi-CLIP, we present additional attention map visualizations in Fig.~\ref{fig_sup1:sup_attention_vis}. As discussed earlier, the visualization indicates that fine-tuning CLIP on a video dataset helps in learning inter-object relationships from temporal cues, which plays a key role in recognizing the action category. Additionally, it steers the models to focus on scene dynamics, moving parts and objects in the scene. For example, in Fig.~\ref{fig_sup1:sup_attention_vis} (left), the model focuses on the moving ball, the child and the fast-moving body parts like the legs. Similarly in Fig.~\ref{fig_sup1:sup_attention_vis} (right), while vanilla CLIP only focuses on the guitar, ViFi-CLIP learns the interaction between the singer and the guitar. 

In Fig.~\ref{fig_sup2:sup_attention_vis_extreme}, we show additional attention map visualizations on extreme out-of-distribution examples. We test ViFi-CLIP on synthetically generated videos to test the generalization ability of the model. Fig.~\ref{fig_sup2:sup_attention_vis_extreme} (left) shows that the models successfully focus on the skating shoes, and the interaction with the teddy-bear. When tested on a rare scenario like `\txt{giraffe tight-roping}' (shown in Fig.~\ref{fig_sup2:sup_attention_vis_extreme} (right)), ViFi-CLIP shows good generalization in recognizing the action using both appearance and temporal cues. These visualizations indicate that temporal relations can be implicitly modeled by simply fine-tuning CLIP on a video-dataset.

\end{document}